\documentclass{article}

\pdfoutput=1
\usepackage[preprint]{arxiv_sty}

\usepackage[utf8]{inputenc} %
\usepackage[T1]{fontenc}    %
\usepackage{hyperref}       %
\usepackage{url}            %
\usepackage{booktabs}       %
\usepackage{amsfonts}       %
\usepackage{nicefrac}       %
\usepackage{microtype}      %
\usepackage{xcolor}         %
\usepackage{graphicx}
\usepackage{algorithm}
\usepackage{color}     %
\usepackage{xcolor}
\usepackage{amsfonts}       %
\usepackage{booktabs}       %
\usepackage{float}
\usepackage{fnpos}
\usepackage{ftnxtra}
\usepackage{graphicx}
\usepackage{microtype}      %
\usepackage{nicefrac}       %
\usepackage{xspace}
\usepackage{mathtools}
\usepackage{thmtools}
\usepackage{colortbl}
\usepackage{wrapfig}
\usepackage{bbm}
\usepackage{tabularx}
\usepackage[capitalize,noabbrev]{cleveref}
\usepackage{wrapfig}
\usepackage{transparent}
\usepackage{enumitem}
\usepackage{changepage} 
\usepackage{pythonhighlight}
\usepackage{tcolorbox}
\usepackage{lipsum} 
\usepackage{booktabs}
\usepackage{multirow}
\usepackage{amsmath}
\usepackage{siunitx}
\usepackage{subfigure}
\usepackage{enumitem}
\usepackage{ulem}

\title{A Human-Like Reasoning Framework for Multi-Phases Planning Task with Large Language Models}

\author{%
  Chengxing Xie \thanks{This work was done when Chengxing Xie was an intern at HKU.} \\
  Department of Computer Science\\
  The University of Hong Kong\\
  \texttt{xiechengxing34@gmail.com} \\  
  \And
  Difan Zou \\
  Department of Computer Science\\
  Institute of Data Science\\
  The University of Hong Kong\\
  \texttt{dzou@cs.hku.hk} \\  
}

\begin{document}

\maketitle

\begin{abstract}

Recent studies have highlighted their proficiency in some simple tasks like writing and coding through various reasoning strategies. However, LLM agents still struggle with tasks that require comprehensive planning, a process that challenges current models and remains a critical research issue. In this study, we concentrate on travel planning, a Multi-Phases planning problem, that involves multiple interconnected stages, such as outlining, information gathering, and planning, often characterized by the need to manage various constraints and uncertainties. Existing reasoning approaches have struggled to effectively address this complex task~\citep{xie2024travelplanner}.
Our research aims to address this challenge by developing a \textbf{human-like planning framework} for LLM agents, i.e., guiding the LLM agent to simulate various steps that humans take when solving Multi-Phases problems. Specifically, we implement several strategies to enable LLM agents to generate a coherent outline for each travel query, mirroring human planning patterns. Additionally, we integrate Strategy Block and Knowledge Block into our framework: Strategy Block facilitates information collection, while Knowledge Block provides essential information for detailed planning. Through our extensive experiments, we demonstrate that our framework significantly improves the planning capabilities of LLM agents, enabling them to tackle the travel planning task with improved efficiency and effectiveness. Our experimental results showcase the exceptional performance of the proposed framework; when combined with GPT-4-Turbo, it attains $10\times$ the performance gains in comparison to the baseline framework deployed on GPT-4-Turbo.

\end{abstract}

\vspace{-5mm}
\section{Introduction}

Recently, large language models (LLMs), such as GPTs \citep{achiam2023gpt4,ouyang2022training_instruct_gpt} and LLaMAs \cite{touvron2023llama}, have demonstrated remarkable potential in handling diverse tasks across various domains, showcasing impressive generalization capabilities. \cite{wang2023avalon} and \cite{mukobi2023welfare} show that LLM agents could play card games with human-like proficiency and, in some instances, even outperform most human players in some games \citep{meta2022human_diplomacy}. Moreover, \cite{zhang2023building}, \cite{dasgupta2023collaborating}, and \cite{fu2024mobile} revealed that LLM agents could solve daily tasks such as cooking and door-opening. Additionally, LLM agents can also establish complex software systems through collaboration with different agent roles \citep{qian2023communicative,huang2023agentcoder}. 

Nevertheless, LLM agents still struggle with certain tasks \cite{wang2023voyager,wu2023smartplay,jimenez2023swe}, which are relatively solvable to humans. 
We categorize these challenges as \textbf{Multi-Phases Planning Tasks}. These tasks are difficult because their solutions can be divided into multiple interconnected phases, each requiring shared information and having dependencies with each other. 
For example, creating a travel plan \citep{xie2024travelplanner} requires humans to plan for multiple days, encompassing various aspects. This involves addressing different elements in different phases of the planning process.
Similarly, solving GitHub issues involves checking various functions and systematically locating and resolving code problems one by one \citep{jimenez2023swe}. The procedures for solving these tasks are often not readily available online and cannot be easily converted into instruction fine-tuning data formats to train LLMs. Consequently, tackling these types of tasks remains a significant challenge for today's LLM agents \citep{wang2023voyager,wu2023smartplay,jimenez2023swe,xie2024travelplanner}.

\begin{figure}[t]
    \centering
    \includegraphics[page=1, width=\textwidth]{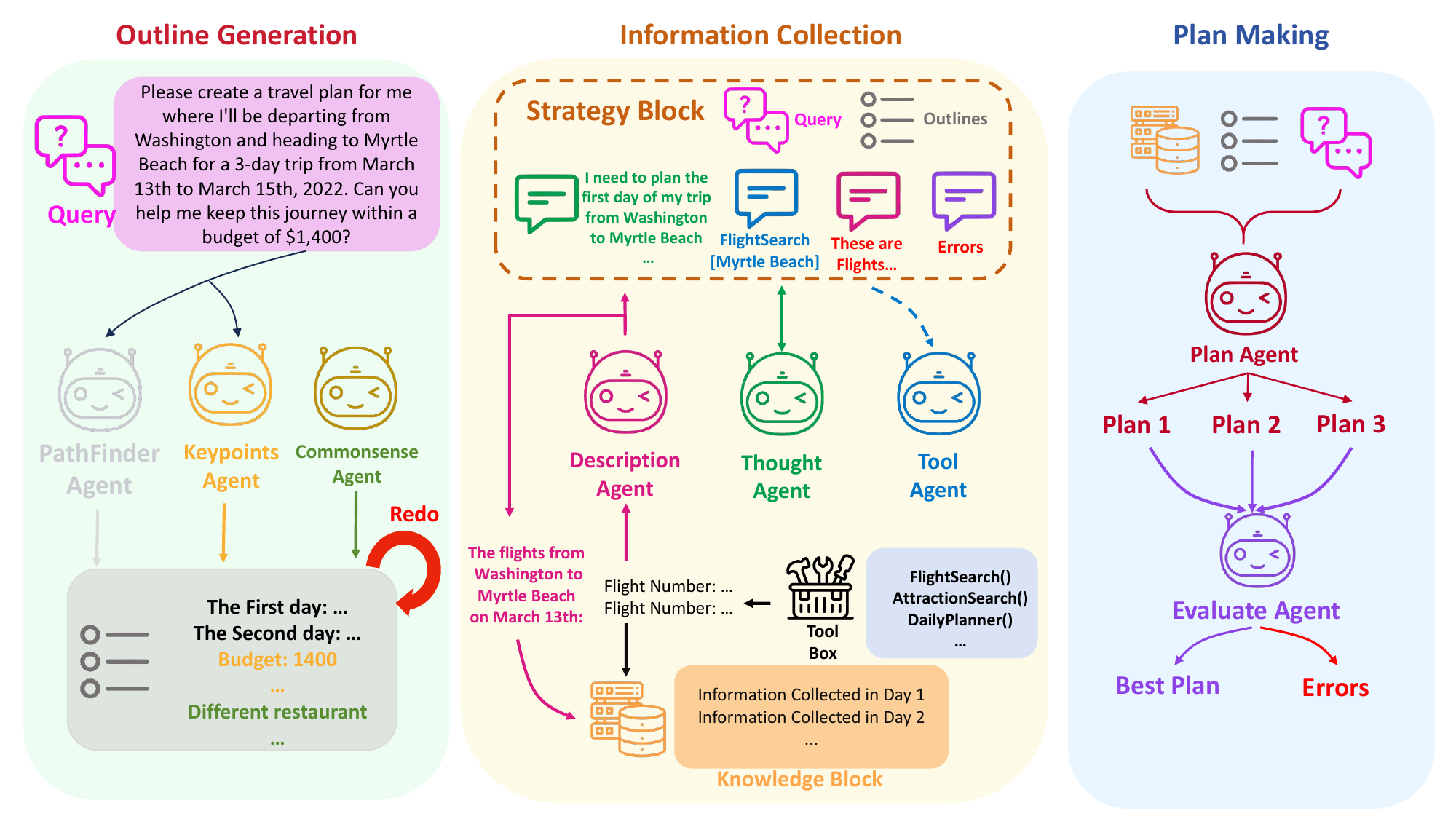}
    \vspace{-4mm}
    \caption{Our human-like planning framework consists of three key parts. In the \textbf{Outline Generation} phase, LLM agents produce rough plans and identify key information related to the query, establishing the foundation for detailed future planning. In the \textbf{Information Collection} phase, LLM agents gather the essential data required for comprehensive planning. Finally, in the \textbf{Plan Making} phase, LLM agents explore the potential plan space and return a well-structured, reasonable plan. }
    \label{fig:framework_main}
    \vspace{-6mm}
\end{figure}

We categorize the stages of solving such tasks into three main phases, detailed as follows:
\textbf{Outline Generation Phase:} When confronted with a Multi-Phase Problem, individuals typically begin by sketching a preliminary outline outlining the path toward a solution. \textbf{Information Collection Phase:} Generally, to solve this type of task, the information provided in the task description alone is completely insufficient. Therefore, individuals must proactively identify the required information and ascertain how to obtain it effectively. \textbf{Plan Making Phase:} Armed with the necessary information, individuals proceed to formulate a plan. Multi-phase problems inherently entail numerous constraints, encompassing factors such as human resources, physical limitations, and more. Consequently, devising effective plans necessitates meticulous consideration of diverse constraints.
Given the challenges posed by Multi-Phase Problems, we have chosen \textbf{travel planning} as our focal point for several reasons. Firstly, it is a time-consuming and often challenging problem for many people, demonstrating its difficulty. Secondly, travel planning inherently involves a multitude of constraints, requires long-term strategizing, and demands significant travel-related information gathering. These complexities make travel planning a fitting subject for exploring and improving planning strategies. 

Previous reasoning strategies \cite{yao2022react, cot, shinn2024reflexion} have shown limited effectiveness in addressing the complexities of travel planning \citep{xie2024travelplanner}. However, by understanding and emulating the processes humans use to tackle Multi-Phases Problems, we can develop more robust planning frameworks to enable LLM agents to tackle such challenges effectively. Our proposed framework aims to capture the essence of human reasoning processes, particularly in the context of Multi-Phase Problems, to enhance the performance of LLM agents.

Our framework is composed of three main components. The first is \textbf{Outline Generation}. When tackling a complex problem, people often create a rough plan to guide them through the process. Similarly, we generate a travel planning outline using multi-agent collaboration, where each agent contributes to a specific part of our travel outline, establishing a foundational guide for planning. The second component is \textbf{Information Collection}. Just as people gather relevant travel information, LLM agents require sufficient data to create reliable plans.
This step ensures that agents have all the necessary details for making informed decisions. 
The third component is \textbf{Plan Making}. After collecting adequate information, the LLM agent creates detailed daily plans. Given the inherent complexity of travel planning, we employ a plan search and evaluation method. Multiple plans are generated in each iteration, and an evaluation agent identifies the best plan and flags errors in each option. Our framework demonstrates that this structured approach leads to impressive performance on this complex task, improving the complex task planning ability of LLM agents. The major findings of our work are summarized as follows:

\begin{itemize}[leftmargin=*]
    \item Our work underscores the challenges LLM agents face with Multi-Phase Planning tasks, highlighting the necessity of employing a novel reasoning framework for effective problem-solving.
    \item We found that utilizing a human reasoning framework for complex tasks is a key factor in enhancing LLM agents' performance on these tasks.
    \item By identifying key factors contributing to effective problem-solving and integrating them into a human-like planning framework, we enable LLM agents to exhibit human-like reasoning, resulting in impressive performance improvements, significantly enhancing LLM agents' performance in travel planning tasks. 
    \item Through extensive experimentation, we validate the effectiveness of our approach, demonstrating its capability to tackle complex planning problems with remarkable success.
\end{itemize}

\section{Related Works}
\subsection{Reasoning Strategy for LLM Agents}
\textbf{Planning without Additional Components:}
In recent times, the development of large language models (LLMs) agents has enabled numerous tasks to be addressed by directly inputting questions into these models. However, LLM agents often struggle with some problem-solving, particularly in areas like mathematics and other intricate planning tasks. To address this limitation, recent research has developed various reasoning strategies aimed at enhancing the problem-solving abilities of LLM agents~\citep{huang2024understanding}. One common strategy, employed by works like~\cite{cot,yao2022react,shen2024hugginggpt}, leverages the divide-and-conquer approach. This involves breaking down a complex task into simpler subtasks and allowing the LLM agents to tackle each subtask sequentially. This method has shown improved performance across a range of tasks. Other works~\citep{cot-sc,yao2024tot,besta2024got,zhao2024large_mcts,xiao2023llm_astar,hao2023reasoning_rap} employ similar strategies using tree searches, such as Monte Carlo Tree Search (MCTS), A*, Breadth-First Search (BFS), and Depth-First Search (DFS). They generate multiple alternative plans via various sampling methods and select the optimal plan through different selection techniques. Reflecting on and refining plans based on prior experiences also significantly improves the planning process of LLM agents~\citep{shinn2024reflexion,gou2023critic,madaan2024self_refine,huang2023recommender}. By reevaluating their plans, LLM agents can avoid recurring errors, thereby enhancing their problem-solving capabilities.

\textbf{Planning with Addition Components:} The planning capabilities of LLM agents can also be improved through interactions with other components. For instance,~\cite{liu2023llm+p},~\cite{guan2023leveraging_ddpl},~\cite{dagan2023dynamic}, and~\cite{cheng2022binding} combine symbolic planners with LLM agents, utilizing the natural language generation capabilities of LLMs to create formalized task descriptions. Other works~\citep{liu2023think,lewis2020retrieval,yang2023coupling,mao2020generation} enhance the problem-solving ability of LLMs by integrating additional memory modules. A particularly effective approach involves enabling LLM agents to access various APIs and tools~\citep{parisi2022talm,qin2023tool,schick2024toolformer,wang2024tools,yang2024gpt4tools,liu2023llava,patil2023gorilla,yuan2024easytool}. Some tools can be directly included in the context~\citep{suris2023vipergpt,shen2024hugginggpt,wu2023visual} as prompt. However, improving an LLM agent’s tool-using abilities often requires generating appropriate datasets and fine-tuning the models~\citep{schick2024toolformer,patil2023gorilla,yang2024gpt4tools}. By providing access to external tools, LLM agents can tackle tasks that are otherwise challenging, such as computational problems and formula verification. Our planning framework integrates these diverse reasoning strategies to improve overall planning capabilities, allowing for more sophisticated problem-solving and strengthening the effectiveness of LLM agents in handling complex tasks.
\subsection{Multi-Agents Framework}

Multi-agent frameworks \citep{guo2024large} have garnered significant interest from researchers due to their flexibility across a broad range of tasks. These frameworks can be utilized in simulations of various kinds, including game simulations \citep{xu2023language, wang2023avalon, mukobi2023welfare}, economic simulations \citep{li2023large, li2023tradinggpt}, and societal simulations \citep{gao2023s, xie2024can, park2023generative, zhao2023competeai, aher2023using}. Moreover, multi-agent frameworks demonstrate superior performance compared to single-agent systems in solving diverse tasks. For instance, \cite{li2023camel}, \cite{hong2023metagpt}, and \cite{chen2023autoagents} have developed general multi-agent frameworks that enhance task performance. For specific tasks, works like \cite{qian2023communicative,hong2023metagpt,huang2023agentcoder} propose specialized frameworks that enable agents to autonomously develop software. In addition, multi-agent systems have been effectively employed for reasoning in embodied environments \cite{dasgupta2023collaborating, zhang2023building} and have demonstrated their potential in scientific research \citep{boiko2023emergent, bran2023chemcrow}. Our framework also incorporates a multi-agent system to enhance overall planning capabilities, leveraging the collaborative and complementary strengths of multiple agents for more effective problem-solving.

\section{Method}

\subsection{Task Description:}

We introduce the task setting of TravelPlanner \cite{xie2024travelplanner}, where the objective is for LLM agents to provide a reasonable travel plan based on a given query, while adhering to the specified constraints and leveraging commonsense knowledge. Each day's plan includes details about the current city, attractions, accommodation, transportation, and dining options (See Appendix \ref{sec:plan_example} for travel plan examples). A reasonable plan must comply with commonsense principles and meet all requirements specified in the travel query, such as avoiding repeated visits to the same attractions and ensuring the plan stays within the specified budget. An initial analysis of GPT-4-Turbo's performance in \cite{xie2024travelplanner} revealed that the most advanced model still struggles with efficient information collection and valid plan making. The details are as follows:
\begin{itemize}[leftmargin=*]
    \item Generating unreasonable travel routes, such as a number of cities visited that is inconsistent with the query or not returning to the original city on the last day.
    \item Utilizing unreasonable transportation methods, like combining self-driving and flight in one trip.
    \item The plan made may include unreal information (hallucination) in the plans.
    \item LLM agents may omit essential information when making the plan and disregard the constraints specified in the query, leading to problematic and incomplete plans.
\end{itemize}

To address the issues of unreasonable travel routes and transportation methods, we employ the \textbf{Outline Generation} phase (Section \ref{sec:outline}) to create logical outlines and appropriate transportation options. To alleviate the hallucination problem, we enhance the \textbf{Information Collection} phase (Section \ref{sec:information_collection}) to ensure more accurate data gathering. Finally, to deal with the missing information during plan creation, we utilize the \textbf{Plan Making} phase (Section \ref{sec:plan_search}) to support the development of correct and comprehensive plans.

\subsection{Framework Overview}

Our framework, as shown in Fig \ref{fig:framework_main}, employs a human-like reasoning approach to address travel planning problems. Based on the specific query, we first generate an outline for the query (Section \ref{sec:outline}). Following the \textit{Outline Generation}, LLM agents equipped with \textit{Strategy block} and \textit{Knowledge Block} proceed with \textit{Information Collection} (Section \ref{sec:information_collection}). During this stage, once sufficient information is gathered for a single day's plan, the agent creates the daily plan in \textit{Plan Making} phase (Section \ref{sec:plan_search}). The final travel plan is composed of each daily plan.

\subsection{Outline Generation}
\label{sec:outline}

Each travel query (details in Appendix \ref{sec:query_example}) involves numerous elements and spans several days, making it challenging to generate a comprehensive plan all at once. When humans tackle such tasks, they typically begin by drafting a rough outline. This outline should encompass the route, transportation options, and key points in the query, facilitating the planning process (details in Appendix \ref{sec:outline_example}). To replicate this approach, we introduce the Route Generation phase, which incorporates transportation evaluation to determine the route and transportation components. Additionally, we integrate two additional agents to generate the key points of the travel, enhancing the usefulness of the outlines.

\begin{figure}[t]
    \centering
    \includegraphics[page=2, width=\textwidth]{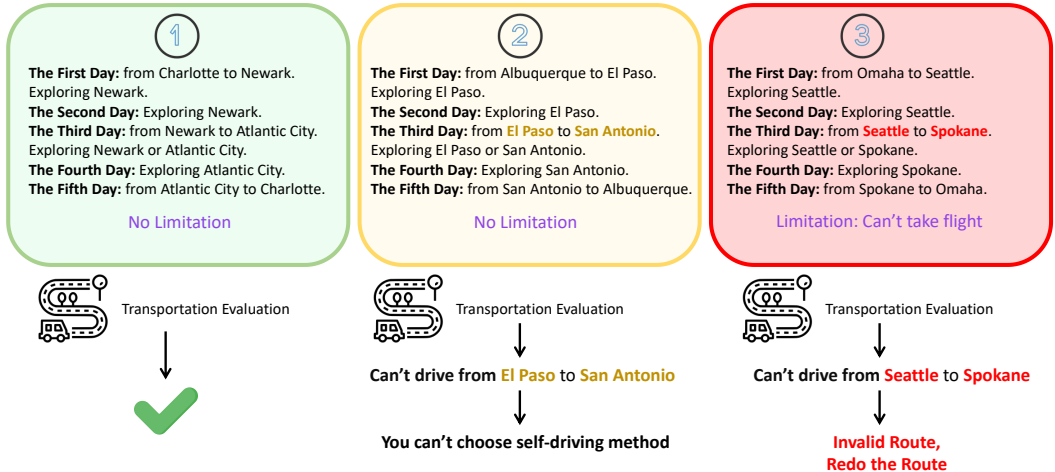}
    \vspace{-5mm}
    \caption{This is our transportation evaluation process for each route generated by PathFinder Agent. We will evaluate whether this route is reasonable under the limitation from the specific travel planning query, e.g., cannot take flight. If not reasonable, we will redo the route by giving feedback. Otherwise, if the route is not reasonable for some kind of transportation, we will add the constraint.}
    \vspace{-5mm}
    \label{fig:route}
\end{figure}

\textbf{Route Generation:}
Drawing inspiration from human reasoning patterns, we've developed the Route Generation component of Outline Generation. Firstly, the query is first passed to the \textbf{PathFinder Agent}. This agent generates a rough route for the entire trip, including city transfers and exploratory travel information. This preliminary route serves as a guide for the subsequent \textit{Information Collection} and \textit{Plan Making} phases, providing a clear roadmap for the journey and making the planning process more structured and transparent.

\textbf{Transportation Evaluation:}
When constructing a travel route, humans typically do not initially verify transportation availability between target cities, assessing options later based on preferences and availability. Similarly, we've observed that LLM agents may generate invalid routes due to a lack of detailed transportation information between cities. To address this issue, we propose adding an evaluation stage after route generation. As shown in Fig \ref{fig:route}, during this stage, each route is evaluated to determine its rationality from a transportation perspective. If the route is perfect, we don't need to do anything. If the route cannot be traveled by driving, we inform the agent accordingly. Similarly, if the route cannot be reached by any means of transportation, we provide the feedback to the PathFinder Agent and prompt the agent to revise the route accordingly.

\textbf{Keypoints Generation:}
When humans receive a query, they focus on the key points and ignore irrelevant details. Inspired by this, we introduced the \textbf{Keypoints Agent} (Fig \ref{fig:framework_main} left) to identify the critical points in the query that need to be considered by the Plan Agent. Moreover, as \cite{xie2024travelplanner} demonstrated, LLM agents tend to overlook common sense when planning, such as navigating the same restaurant or attraction. To address this issue, the \textbf{Commonsense Agent} (Fig \ref{fig:framework_main} left) generates basic guides related to travel planning. This agent \textbf{doesn't} have access to the specific query, as we believe these guides are common sense and universally applicable to every query.

These three parts combine with each other to generate a useful and detailed travel outline, building a strong foundation for the planning task.
\vspace{-2mm}
\subsection{Information Collection}
\label{sec:information_collection}
\vspace{-2mm}

After the \textbf{Outline Generation} phase, humans proceed to collect necessary travel-related information, such as specific attractions and restaurants.
Collecting sufficient information is crucial as it can significantly decrease the probability of hallucination. In our framework, the Information Collection process begins with the \textbf{Thought Agent} generating the next steps based on the Strategy Block. Subsequently, the \textbf{Tool Agent} utilizes the output of the Thought Agent to generate a suitable function expression. The result of this function, primarily comprising travel information, is then recorded in the Knowledge Block with a description from the \textbf{Description Agent}. Finally, the selected information is forwarded to the Plan Agent for generating the daily travel plan.

\textbf{Strategy Block:}
In the information collection process, humans typically remember the types of information they have gathered, which guides their subsequent steps. To emulate this procedure, we introduce a component called the \textbf{Strategy Block} as shown in Fig \ref{fig:framework_main}, primarily used by the \textbf{Thought Agent}. First, the outline is stored in the Strategy Block to guide the information collection process. The Strategy Block also informs the Thought Agent which day it is in the travel plan and short descriptions of the collected data. Besides, we streamline the prompts for better workflow management. Compared to the scratchpad approach in \cite{xie2024travelplanner}, our design provides more task-specific details and better supports the Thought Agent in collecting information.

\textbf{Tool Agent:} We've observed that as the planning process progresses, the context length can become too long, leading to potential oversight of tool documents placed at the beginning of input messages. Thus, we introduce the Tool Agent tasked with generating the correct function expression format messages. The Tool Agent's input includes only the tool document, the query, and some previous content from the Strategy Block to ensure it can generate the correct information based on the context. 

\begin{wrapfigure}{t}{0.5\textwidth}
\vspace{-5mm}
  \centering
    \includegraphics[page=1,width=0.5\textwidth]{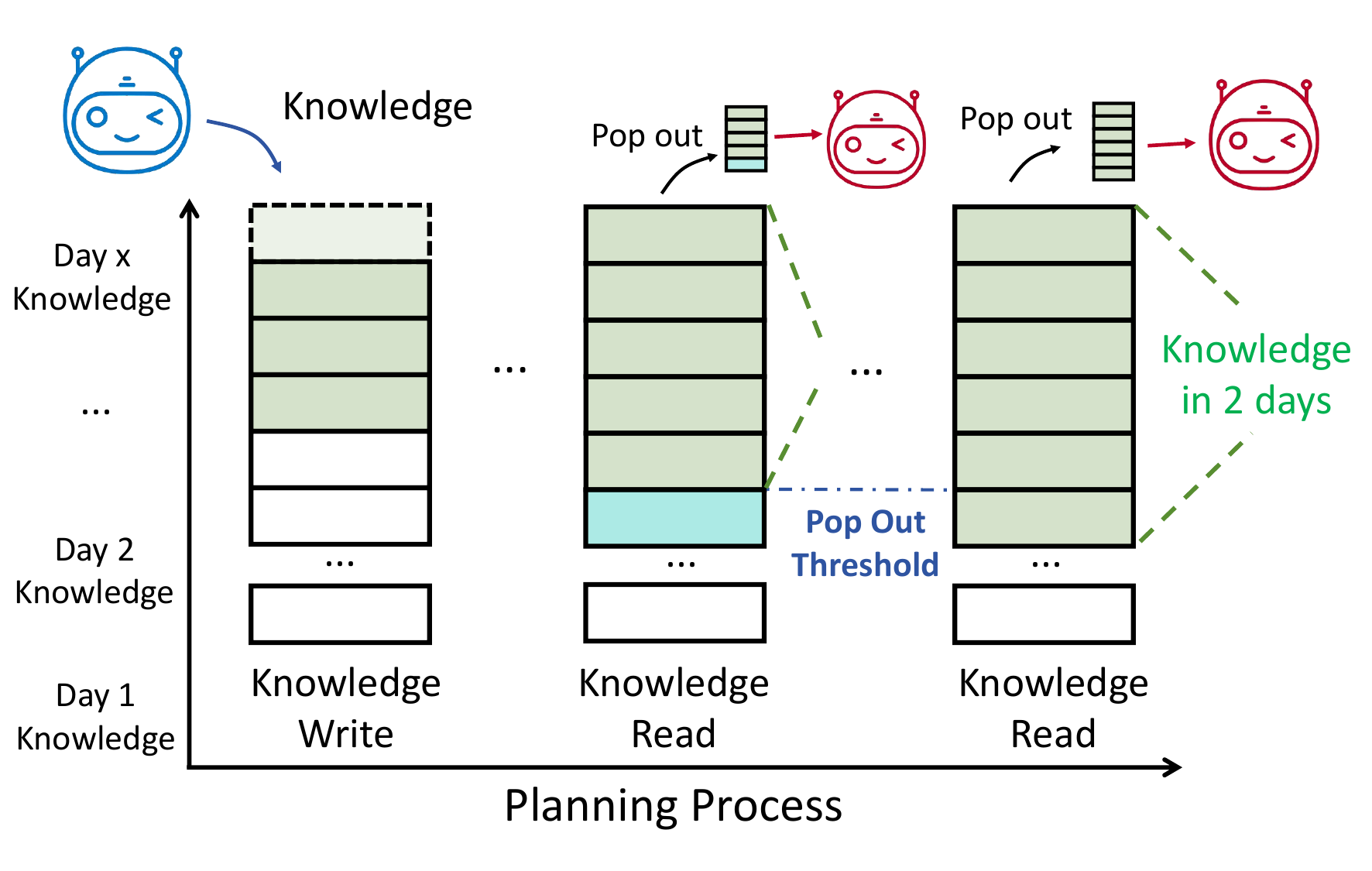}
    \vspace{-6mm}
\caption{This image illustrates the Knowledge Block workflow. The left part is the \textbf{Knowledge Write} process, where a function's result is written to the top of the Knowledge Block. About \textbf{Knowledge Read}, when the knowledge needs to be popped out but is below the threshold, the Knowledge Block pops out enough items to meet the threshold. If the number exceeds the threshold, the items in past two days are popped out.}
  \label{fig:knowledge_block}
  \vspace{-5mm}
\end{wrapfigure}

\textbf{Knowledge Block:}
Humans will make their plans by examining the detailed information they have previously collected. To mimic this process, we introduce the \textbf{Knowledge Block}, as shown in Fig \ref{fig:framework_main}, primarily used by the \textbf{Plan Agent}. In Fig \ref{fig:knowledge_block} left part, this block will automatically record all information and the description of each information, which is generated by \textbf{Description Agent}. However, when travel plans span long periods, the extensive information can make it difficult to extract pertinent details and may even exceed the LLM's context window length. To address this, we implement a stack-like structure within the Knowledge Block. When the Plan Agent needs detailed information, the block only ``pops out'' the information collected over the past two days, restricted by a minimal threshold. As shown in Fig \ref{fig:knowledge_block} If the number of knowledge items collected exceeds the threshold, the block pops out the same number of items. If it exceeds the threshold, all knowledge items from the past two days are popped out.

Lastly, we found that presenting the data in a DataFrame format makes it difficult to read and interpret. Therefore, we have reorganized the information into a more reader-friendly structure. We directly specify each data's name just before presenting the data, like: \texttt{Flight Number: F3502691} (see Appendix \ref{sec:knoledge_block_format} for more details).

\vspace{-2mm}
\subsection{Plan Making}
\label{sec:plan_search}
\vspace{-2mm}

When humans make plans, they typically do not plan everything at once; instead, they approach it step by step. Therefore, unlike the approach in \cite{xie2024travelplanner} that plans the entire trip at once, we adopt a daily planning strategy. This method requires less information at each step, making the planning process easier. Each time the Thought Agent determines that sufficient information has been collected for a specific day, the Tool Agent calls the \textit{DailyPlanner} tool, and the Plan Agent creates the daily plan.

\textbf{Plan Search:} 
When examining the results of Plan Agent, we observe that creating a plan often introduces various errors. These errors can stem from insufficient information collected during the information collection stage or from oversights during the planning stage. To address this issue, we propose a plan search method.

Each time the Plan Agent generates a daily plan, it creates several plans based on the same information. An \textbf{Evaluate Agent} reviews each plan, converting them into JSON format and using code to identify and rank errors. Based on these evaluations, we select the best plan. If significant errors are found, these errors are recorded in the Strategy Block, prompting further information collection, and the plans with errors are discarded. This discard process occurs only once per day's plan to prevent unlimited planning iterations. Through daily planning and plan search strategies, the agent can generate more reasonable and reliable travel plans.

\vspace{-2mm}
\section{Experiment}
\vspace{-1mm}
\definecolor{CustomColor1}{HTML}{c1f1fc}
\definecolor{CustomColor2}{HTML}{ebffac}
\definecolor{CustomColor3}{HTML}{ffc2e5}

\newcolumntype{a}{>{\columncolor{CustomColor1}}c}
\newcolumntype{b}{>{\columncolor{CustomColor2}}c}
\newcolumntype{d}{>{\columncolor{CustomColor3}}c}
\setlength\tabcolsep{2pt}

\begin{table}[t]
\centering
\caption{Our experimental results on the validation dataset of the TravelPlanner benchmark demonstrate that our framework enables GPT-3.5 to surpass previous GPT-4's performance. Additionally, our framework increases the Final Pass Rate of GPT-4-Turbo $10\times$ compared to previous algorithms. For each model, using our framework results in significantly improved performance.}
\vskip -.1in
\resizebox{1.0\linewidth}{!}{
\begin{tabular}{>{\centering\arraybackslash}m{3cm}cccccc}
\toprule

\textbf{ \vspace{-1mm} Model} & \multirow{2}{*}{\begin{tabular}[c]{@{}c@{}}Delivery Rate\end{tabular}} & \multicolumn{2}{c}{\begin{tabular}[c]{@{}c@{}}Commonsense Pass Rate\end{tabular}} & \multicolumn{2}{c}{\begin{tabular}[c]{@{}c@{}}Hard Constraint Pass Rate\end{tabular}} & \multirow{2}{*}{\begin{tabular}[c]{@{}c@{}}Final Pass Rate\end{tabular}} \\ \cmidrule(l){3-4} \cmidrule(l){5-6}
         &  & Micro & Macro & Micro   & Macro  &     \\ \midrule

\rowcolor[gray]{0.85}
\multicolumn{7}{c}{\textbf{\textit{TravelPlanner Result}}} \\ \midrule

\rowcolor{CustomColor2}
\multicolumn{1}{l}{\cellcolor{white}Mistral-7B-32K~\cite{jiang2023mistral}} & \num{8.9} & \num{5.9} & \num{0} & \num{0} & \num{0} & \num{0} \\  

\rowcolor{CustomColor2}
\multicolumn{1}{l}{\cellcolor{white}Mixtral-8×7B-MoE~\cite{jiang2024mixtral}} & \num{49.4} & \num{30.0} & \num{0} & \num{1.2} & \num{0.6} & \num{0} \\

\rowcolor{CustomColor2}
\multicolumn{1}{l}{\cellcolor{white}GPT-3.5-Turbo~\cite{ouyang2022training_instruct_gpt}} & \num{86.7} & \num{54.0} & \num{0} & \num{0} & \num{0} & \num{0} \\

\rowcolor{CustomColor2}
\multicolumn{1}{l}{\cellcolor{white}GPT-4-Turbo~\cite{achiam2023gpt4}} & 89.4 & 61.1 & 2.8 & 15.2 & 10.6 & 0.6 \\

\midrule
\rowcolor[gray]{0.85}
\multicolumn{7}{c}{\textbf{\textit{Our Results}}} \\ \midrule

\rowcolor{CustomColor1}
\multicolumn{1}{l}{\cellcolor{white}Mistral-7B-32K~\cite{jiang2023mistral}} & \num{39.4} & \num{24.0} & \num{1.1} & \num{0.5} & \num{0.6} & \num{0.6} \\  

\rowcolor{CustomColor1}
\multicolumn{1}{l}{\cellcolor{white}Mixtral-8×7B-MoE~\cite{jiang2024mixtral}} & \num{67.8} & \num{40.2} & \num{0} & \num{0} & \num{0} & \num{0} \\

\rowcolor{CustomColor1}
\multicolumn{1}{l}{\cellcolor{white}GPT-3.5-Turbo~\cite{ouyang2022training_instruct_gpt}} & \textbf{100.0} & \textbf{75.1} & \num{15.6} & \num{15.5} & \num{4.4} & \num{2.2} \\

\rowcolor{CustomColor1}
\multicolumn{1}{l}{\cellcolor{white}GPT-4-Turbo~\cite{achiam2023gpt4}} & \num{91.7} & \num{74.6} & \textbf{24.4} & \textbf{35.7} & \textbf{16.7} & \textbf{6.7} \\

\bottomrule
\end{tabular}}
\label{tab:main_res}
\vspace{-.1cm}
\end{table}

\subsection{Experiment Setup}
\label{sec:exp_set_up}
We utilize the TravelPlanner benchmark proposed by \cite{xie2024travelplanner}, which includes a variety of travel queries with different travel lengths and difficulty levels. This benchmark has a train, valid, and test set. The valid set has 180 queries and the test set has 1000 queries. We use the valid set to examine our framework as running the experiment on the test set is too expensive. We use gpt-3.5-turbo-1106, gpt-4-1106-preview, mixtral, mistral-7B-32K in our experiments \cite{achiam2023gpt4,jiang2024mixtral,jiang2023mistral,ouyang2022training_instruct_gpt}. We ran open-source models on 4 NVIDIA RTX A6000.

\textbf{Hyperparameters Settings:}
To ensure the reproducibility of our experimental results, we set the temperature to 0 for all processes except for Plan Agent, where it is set to 0.7 to allow the Plan Agent to generate diverse plans enabling a larger search space.

For the travel route generation, we limit the maximum number of retries to 3 to prevent infinite remaking of the travel route. The Plan Agent generates 3 different plans each time. If all generated plans contain errors, we revert to the information collection stage only once. If errors persist after this second attempt, we proceed with the process regardless. About the Knowledge Block, we set the minimal number of pop-outs is 5.

\textbf{Metrics:}
We use the same metrics proposed in TravelPlanner \cite{xie2024travelplanner}.
\begin{itemize}[leftmargin=*]
\item \textbf{Delivery Rate:} Measures if agents can successfully deliver a final plan within a set number of steps (max 45). Failure includes dead loops, numerous failed attempts, or exceeding the step limit.
\item \textbf{Commonsense Constraint Pass Rate:} Assesses if agents can incorporate commonsense into their plans without explicit instructions, across eight dimensions.
\item \textbf{Hard Constraint Pass Rate:} Evaluates if a plan meets all explicitly given hard constraints, testing agents' adaptability to diverse user queries.
\item \textbf{Final Pass Rate:} Indicates the proportion of plans that meet all constraints (delivery, commonsense, and hard constraints), reflecting agents' overall proficiency in producing practical plans.

\item \textbf{Micro Pass Rate:} The Micro Pass Rate evaluates the proportion of constraints that are successfully passed by an agent's plans, as shown in Formula \ref{formula:metric}. It is calculated by taking the total number of successfully met constraints across all plans and dividing it by the total number of constraints applied to all plans.
\item \textbf{Macro Pass Rate:} The Macro Pass Rate assesses the proportion of plans that satisfy all of their constraints. It calculates the ratio of the number of plans that meet all applicable commonsense or hard constraints to the total number of plans evaluated.
\begin{equation}
     \quad \text{Micro Pass Rate} = \frac{\sum_{p \in P} \sum_{c \in C_p} 1_{\text{passed}(c, p)}}{\sum_{p \in P} |C_p|}, \text{Macro Pass Rate} = \frac{\sum_{p \in P} 1_{\text{passed}(C_p, p)}}{|P|} 
     \label{formula:metric}
\end{equation}

\( P \) is the set of all plans. \( C_p \) is the set of constraints applicable to a specific plan \( p \). \( 1_{\text{passed}(c, p)} \) is an indicator function that returns 1 if constraint \( c \) is passed in plan \( p \), and 0 otherwise. \( 1_{\text{passed}(C_p, p)} \) is an indicator function that returns 1 if all constraints in \( C_p \) are passed in plan \( p \), and 0 otherwise.
\end{itemize}

\subsection{Experiment Result}

\begin{wrapfigure}{t}{0.5\textwidth}
\vspace{-8mm}
  \centering
    \includegraphics[width=0.4\textwidth]{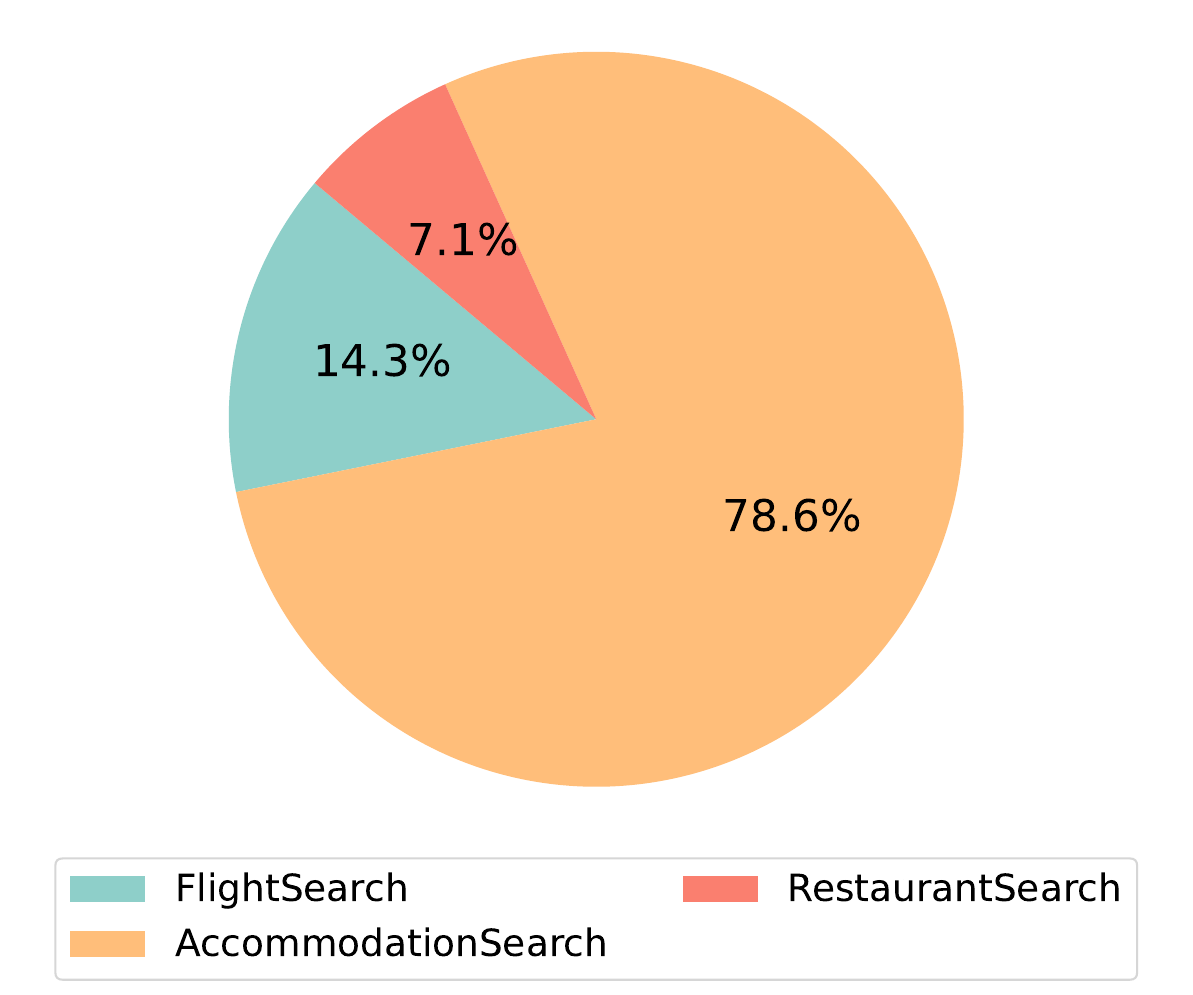}
    \vspace{-3mm}
\caption{The error distribution about GPT-4-Turbo's Delivery Rate Failure.}. 

  \label{fig:delivery}
  \vspace{-6mm}
\end{wrapfigure}

As shown in Table \ref{tab:main_res}, we tested four LLMs on the dataset to verify our framework's effectiveness. All models exhibited increased Delivery Rate, with Mistral-7B showing an improvement of 8.9\% to 39.4\% and GPT-3.5-Turbo achieving 100\% Delivery Rate. This indicates that GPT-3.5-Turbo can generate a travel plan for every query, strongly demonstrating that our framework enables better planning for the task.

However, we observed limited improvement for GPT-4-Turbo on Delivery Rate. To understand the reason, we visualized the error distribution from GPT-4-Turbo in Fig \ref{fig:delivery}. We found that the primary issue was the repeated use of the same function three times, with the most frequent being AccommodationSearch. The agent repeatedly attempted to find the 'correct' room type, often overlooking some information, which led to redundancy. Additionally, GPT-4-Turbo aimed to find accommodations that could fit all travelers in one room, ignoring the possibility of booking multiple rooms. The lower Delivery Rate will cause a lower Micro Commonsense pass rate. We conjecture that GPT-4-Turbo's stronger reasoning capabilities hindered its progress by overcomplicating the task.

\begin{figure}[!tbp]
	\centering
        \subfigure{\includegraphics[width=.52\textwidth]{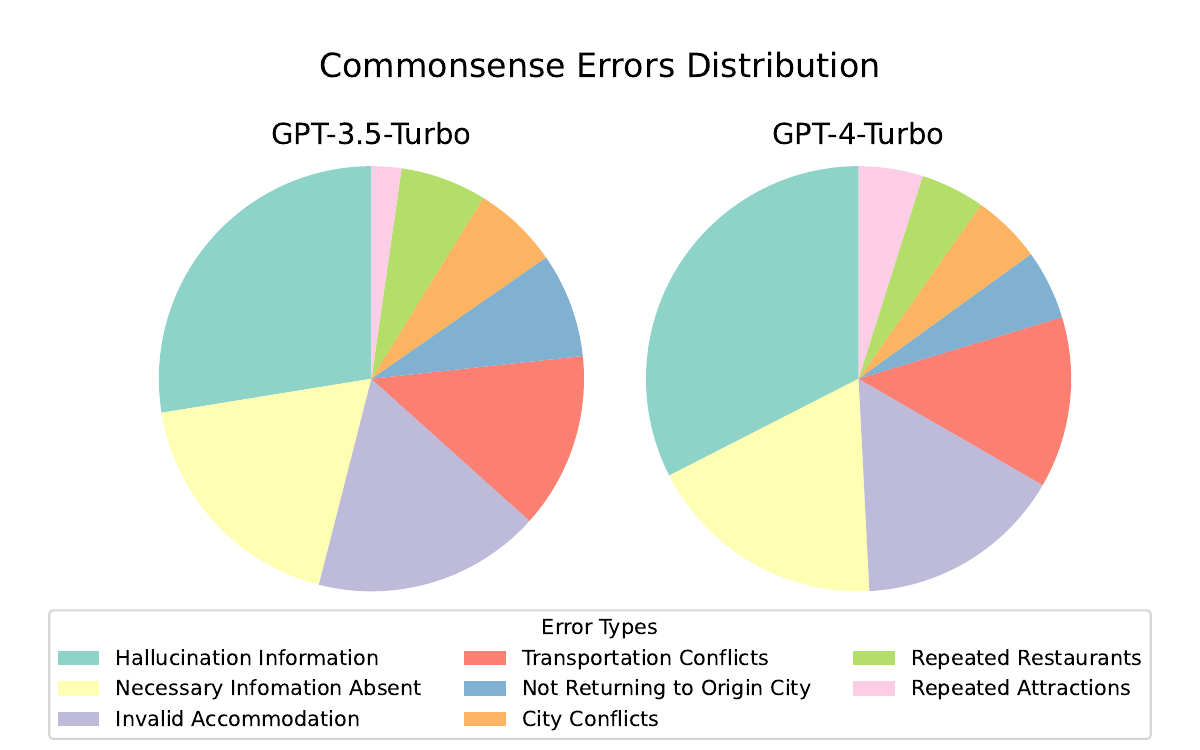}}
	\subfigure{\includegraphics[width=.46\textwidth]{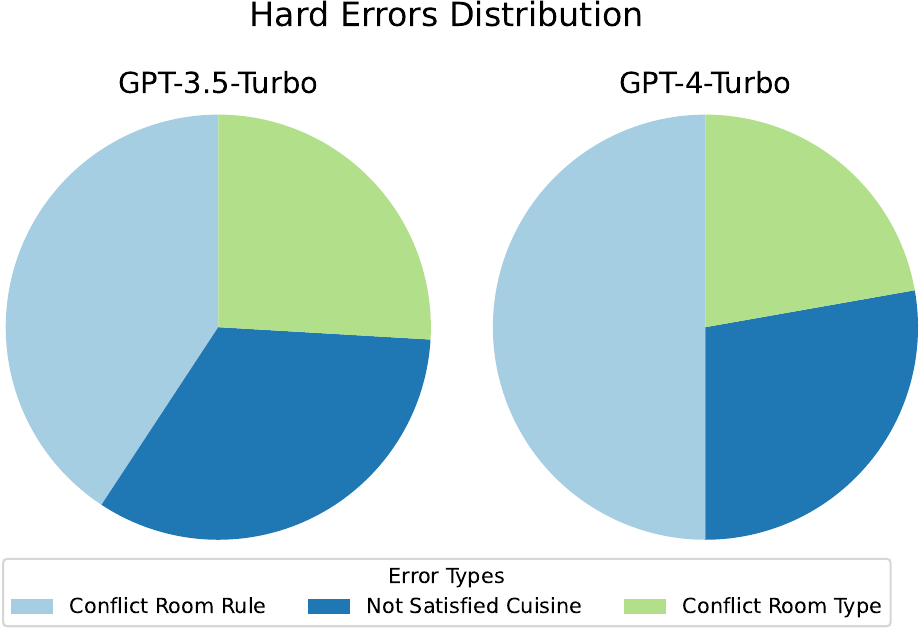}}
 \vskip -.1in
	\caption{The error distribution of GPT-3.5-Turbo and GPT-4-Turbo.}
	\label{fig:constraints_error}
\end{figure}

\textbf{Commonsense Pass Rate:} Our framework demonstrates impressive improvements in the Commonsense Pass Rate across all models. For GPT-3.5-Turbo, the Micro Pass Rate increased from 54.0\% to 75.1\%, and the Macro Pass Rate increased from 0\% to 15.6\%. For GPT-4-Turbo, the Micro Pass Rate increased from 61.1\% to 74.6\%, while the Macro Pass Rate saw a significant boost from 2.8\% to 24.4\%. Although Mistral and Mixtral showed improvements in their Micro Pass Rates, their Macro Pass Rates did not see significant increases.

Analyzing the commonsense error distribution in Fig \ref{fig:constraints_error}, we find that GPT-3.5-Turbo and GPT-4-Turbo exhibit very similar error patterns. The top three errors are Hallucinated Information, Necessary Information Absent, and Invalid Accommodation, accounting for nearly 60\% of all errors. Hallucinated Information indicates that the LLM agents still generate unreal information. Necessary Information Absent suggests that the Plan Agent might not strictly follow instructions or the information collection phase misses key details. Invalid Accommodation implies that some points in the accommodation data are overlooked.

\textbf{Hard Constraints Pass Rate: } Additionally, our framework enhances the Pass Rate for Hard Constraints. For GPT-4-Turbo, the Micro Pass Rate increased from 15.2\% to 35.7\%, and the Macro Pass Rate increased from 10.6\% to 16.7\%. While this improvement is not as substantial as that of the Commonsense Pass Rate, it is still a notable enhancement. For GPT-3.5-Turbo, the improvements were slightly lower, with increases from 0\% to 15.5\% for the Micro Pass Rate and from 0\% to 4.4\% for the Macro Pass Rate. The last two models, Mistral and Mixtral, showed only slight improvements. Examining the error distribution in Fig \ref{fig:constraints_error}, we observe that the main errors mainly appear in accommodation planning. This suggests that the concentrated information about each accommodation is challenging for LLM agents to correctly extract in making accurate plans.

Regarding the Final Pass Rate, GPT-3.5-Turbo's result using our framework surpasses the GPT-4-Turbo's result in the previous algorithm, showcasing the effectiveness of our framework. Notably, GPT-4-Turbo's Final Pass Rate increased $10\times$ compared to the previous algorithm. This significant improvement underscores our framework's capability to enhance the performance and reliability of LLM agents in complex travel planning tasks.
\subsection{Ablation Study}

\begin{table}[t]
\centering
\caption{Ablation study results on GPT-3.5-Turbo.}
\vskip -.1in
\resizebox{1.0\linewidth}{!}{
\begin{tabular}{lcccccc}
\toprule
\tiny
         
         & \multirow{2}{*}{\begin{tabular}[c]{@{}c@{}}Delivery Rate\end{tabular}} & \multicolumn{2}{c}{\begin{tabular}[c]{@{}c@{}}Commonsense Pass Rate\end{tabular}} & \multicolumn{2}{c}{\begin{tabular}[c]{@{}c@{}}Hard Constraint Pass Rate\end{tabular}} & \multirow{2}{*}{\begin{tabular}[c]{@{}c@{}}Final Pass Rate\end{tabular}} \\ \cmidrule(l){3-4} \cmidrule(l){5-6}
         &  & Micro & Macro & Micro   & Macro  &     \\ \midrule

\rowcolor{CustomColor2}
\multicolumn{1}{l}{\cellcolor{white}Original Result \cite{xie2024travelplanner}} & \num{86.7} & \num{54.0} & \num{0} & \num{0} & \num{0} & \num{0} \\

\rowcolor{CustomColor3}
\multicolumn{1}{l}{\cellcolor{white}No Outlines} & \num{93.3} & \num{59.2} & \num{1.1} & \num{3.8} & \num{1.1} & \num{0} \\

\rowcolor{CustomColor3}
\multicolumn{1}{l}{\cellcolor{white}No Strategy} & \num{100} & \num{74.9} & \num{13.9} & \num{12.1} & \num{2.2} & \num{0.6} \\

\rowcolor{CustomColor3}
\multicolumn{1}{l}{\cellcolor{white}No Knowledge} & \num{99.4} & \num{66.0} & \num{8.3} & \num{10.2} & \num{1.1} & \num{0} \\

\rowcolor{CustomColor3}
\multicolumn{1}{l}{\cellcolor{white}No Plan Search} & \num{99.4} & \num{67.4} & \num{0.6} & \num{1.9} & \num{0} & \num{0} \\

\rowcolor{CustomColor1}
\multicolumn{1}{l}{\cellcolor{white}Our Framework} & \textbf{100.0} & \textbf{75.1} & \textbf{15.6} & \textbf{15.5} & \textbf{4.4} & \textbf{2.2} \\

\bottomrule
\vspace{-4mm}
\end{tabular}}
\label{tab:ablation}
\end{table}

We analyzed the impact of each component of our framework by removing them one at a time, as shown in Table \ref{tab:ablation}. When the outline generation component was removed, we observed a significant drop in all metrics, although they remained better than the TravelPlanner results. This indicates that Outlines Generation is a crucial element of our framework.

About Strategy Block, we removed some information from the Strategy Block, such as specific daytime guidance, Knowledge Block Information, Budget Requirements, and key points from the query. However, the rest of this block remains necessary because, without it, the information collection process cannot proceed. From the results, we observed that all metrics decreased. However, the degree of decrease in Delivery Rate and Micro Commonsense Pass Rate was not very substantial. This outcome is expected, as this component does not significantly impact the feasibility of creating a plan but rather the quality of the plans produced.

For the Knowledge Block, removing it entirely is impractical. Therefore, we adopted the method from \cite{xie2024travelplanner}, using an extra tool to record information. To prevent exceeding the maximum token limit, we implemented a maximum length limitation. 
Removing this block caused a notable decline in all metrics except Delivery Rate, echoing findings from the Strategy Block ablation study. These outcomes underscore the critical role of the Knowledge Block.

Additionally, when we removed the Plan Search component, we observed a dramatic drop in all metrics except Delivery Rate. The decrease was particularly severe in the Macro Commonsense Pass Rate compared to other ablation studies, highlighting the critical role of the Plan Search component.

Our ablation study results confirm that each block of our framework is essential. Removing any part will significantly decrease performance across various metrics. These parts are vital for managing information and enhancing plan quality, ensuring a robust and efficient planning process.

\vspace{-2mm}
\section{Conclusion and Future Work}
\label{sec:conclution_and_limitation}
\vspace{-2mm}

In this paper, we presented a novel human-like reasoning framework for travel planning using LLM Agents. Our approach integrates several key components: Outline Generation, Information Collection with blocks, and Plan Search. These components synergistically mimic human problem-solving strategies, enabling LLM agents to handle multi-phase tasks more effectively. Extensive experiments on the TravelPlanner benchmark demonstrated significant improvements across multiple models with our framework.

Future work can focus on applying our planning framework to real-world planning problems in practical environments. Additionally, the current tools within our framework are limited. Enhancing the framework by incorporating more advanced tools, such as those with additional parameters and capabilities, could further improve its planning and problem-solving effectiveness. This expansion could enable more sophisticated and adaptable solutions, making our framework even more robust in addressing complex, real-world challenges.

\newpage

\bibliography{nips}
\bibliographystyle{plain}

\appendix
\newpage
\section{Appendix / supplemental material}

\subsection{Limitation}
\label{limitation}

Although our framework has significantly improved performance compared to previous studies, achieving a Final Pass Rate of 100\% akin to human performance remains elusive. One limitation we observed is that even when provided with detailed information, LLM Agents still tend to overlook certain key points. Addressing this challenge may require the integration of alternative planning strategies. Additionally, effectively utilizing extensive accommodation-related information to generate reasonable plans remains a significant challenge, even for advanced models like GPT-4-Turbo. Furthermore, the occurrence of hallucinations during the planning process presents another obstacle that requires attention.

In future work, addressing these limitations could involve exploring novel planning strategies that complement the capabilities of LLM Agents, such as incorporating heuristics or domain-specific rules. Moreover, developing more sophisticated models that can better comprehend and utilize complex accommodation-related information could lead to more accurate and reliable travel plans. Additionally, investigating techniques to mitigate the occurrence of hallucinations and improve the overall robustness of LLM-based planning systems could further enhance their performance. Overall, overcoming these challenges will be crucial for advancing the state-of-the-art in LLM-driven travel planning systems.

\subsection{Impact}
\label{sec:impact}
The social impact of our work is profound, as it pioneers a novel methodology by emulating human reasoning processes to address Multi-Phases planning challenges. Equipping LLM agents with human-like problem-solving abilities, our framework enhances their efficacy in applications, such as travel planning, thereby potentially revolutionizing various industries reliant on complex decision-making. However, the advent of LLM agents with enhanced planning capabilities also raises concerns about potential job displacement and ethical implications surrounding data privacy and algorithmic bias, necessitating careful consideration and regulation to mitigate adverse consequences.

\subsection{Ethics Statement}
\label{sec:ethics}
In our study, we introduce a novel human-like reasoning framework to enhance LLM agents' proficiency in Multi-Phases planning tasks, especially on traveling planning problems. While our work aims to advance the capabilities of LLM agents for beneficial purposes, we recognize the potential for misuse or unintended consequences, such as the propagation of inaccurate information. To address these ethical considerations, we advocate for ongoing research into methods for detecting and mitigating biases in LLMs, as well as the promotion of responsible use and transparency in the deployment of such frameworks. We urge users and researchers to remain vigilant of these ethical risks and prioritize the ethical considerations in the development and application of our framework.

\clearpage
\newpage
\subsection{Example}

\subsection{Travel Query Example:}
\label{sec:query_example}

\begin{wrapfigure}{t}{\textwidth}
\begin{tcolorbox}[halign=left, width=\linewidth, boxrule=3pt, colback=gray!20, colframe=gray!20, left=0pt, right=10pt]
\fontfamily{pcr}\selectfont
1. Could you please create a 5-day travel itinerary for one person, starting in Albuquerque and visiting 2 cities in Texas from March 25th to March 29th, 2022? The travel plan should work within a budget of \$2,100.

2. Could you help create a 7-day travel plan for a group of 3, departing from Greensboro and touring 3 different cities in Georgia from March 10th to March 16th, 2022? We have a new budget of \$4,000 for this trip. We'd also appreciate if our accommodations have smoking areas.

3. Could you help create a 5-day itinerary for a travel plan departing from Grand Junction and heading to 2 cities in Arizona from March 19th to March 23rd, 2022? It's a plan for two people with a budget of \$2,100. Our accommodations should allow visitors and our preference is for private rooms. Additionally, we do not require any flight transportation.

4. Could you create a 7-day travel itinerary for 2 people, departing from Albuquerque and visiting 3 cities in Texas from March 8th to March 14th, 2022? Our budget is set at \$5,000. We require accommodations that allow smoking and are preferably not shared rooms. We would prefer to avoid any flights for our transportation.
\end{tcolorbox}
\vspace{-3mm}
\caption{Few examples of the travel query}
\label{fig:travel_query}
\end{wrapfigure}

\clearpage
\newpage

\subsection{Travel Plan Example:}
\label{sec:plan_example}

\begin{wrapfigure}{t}{\textwidth}
\begin{tcolorbox}[halign=left, width=\linewidth, boxrule=3pt, colback=gray!20, colframe=gray!20, left=0pt, right=10pt]
\fontfamily{pcr}\selectfont
"day": 1,

"current\_city": "from Buffalo to Atlanta",

"transportation": "Flight Number: F3502691, from Buffalo to Atlanta, Departure Time: 18:48, Arrival Time: 20:51",

"breakfast": "-",

"attraction": "Georgia Aquarium, Atlanta;World of Coca-Cola, Atlanta;",

"lunch": "-",

"dinner": "Chaina Ram Sindhi Confectioners, Atlanta",

"accommodation": "Spacious private room close St. Barnabas Hospital, Atlanta"

"day": 2,

"current\_city": "Atlanta",

"transportation": "-",

"breakfast": "Baba Au Rhum, Atlanta",

"attraction": "Atlanta Botanical Garden, Atlanta;High Museum of Art, Atlanta;",

"lunch": "Barkat, Atlanta",

"dinner": "Taste of Vishal, Atlanta",

"accommodation": "Spacious private room close St. Barnabas Hospital, Atlanta"

"day": 3,

"current\_city": "from Atlanta to Buffalo",

"transportation": "Flight Number: F3502694, from Atlanta to Buffalo, Departure Time: 15:47, Arrival Time: 17:42",

"breakfast": "Asian Bistro, Atlanta",

"attraction": "Piedmont Park, Atlanta;",

"lunch": "Beliram Degchiwala, Atlanta",

"dinner": "-",

"accommodation": "-"

\end{tcolorbox}
\vspace{-3mm}
\caption{An example of the travel plan}
\label{fig:travel_plan_example}
\end{wrapfigure}

\clearpage
\newpage

\subsection{Outline Example:}
\label{sec:outline_example}

\begin{wrapfigure}{t}{\textwidth}
\begin{tcolorbox}[halign=left, width=\linewidth, boxrule=3pt, colback=gray!20, colframe=gray!20, left=0pt, right=10pt]

\fontfamily{pcr}\selectfont

The First Day: from Buffalo to Atlanta. Exploring Atlanta.

The Second Day: Exploring Atlanta.

The Third Day: from Atlanta to Buffalo.

1.Departure and Return Dates: The travel must commence on March 2nd, 2022, and conclude with a return to Buffalo on March 4th, 2022.

2.Solo Travel: The plan should be tailored for a single traveler, ensuring accommodations and activities are suitable for one person.

3.Budget Limit: The total cost of the trip, including transportation, accommodation, meals, and activities, must not exceed \$1,100.

4.Destination Specific: All travel arrangements must be made with Atlanta as the sole destination for the duration of the trip.

5.Time Frame: The entire travel plan, including all experiences and bookings, must fit within the 3-day window specified.

1. Accommodations should be booked for the correct dates corresponding to the traveler's stay in each city, ensuring no gaps or overlaps in lodging.

2. Dining options for breakfast, lunch, and dinner should be reasonably spaced throughout the day and geographically convenient to either the accommodation or planned attractions.

3. Transportation arrangements must align with the traveler's schedule, allowing sufficient time for check-in, security clearance at airports, and transfers between different modes of transport.

4. Attractions should be planned with consideration of their operating hours and any required advance reservations or tickets, avoiding scheduling conflicts.

5. The sequence of visiting attractions should be logical and efficient, minimizing backtracking and excessive travel time between locations.

6. Ensure that the transportation method selected is appropriate for the distance and time required to reach the scheduled activities, such as not planning long-distance road trips between breakfast and lunch.

7. Dining reservations should be made with consideration of dietary restrictions, restaurant capacity, and the expected duration of the meal to maintain the day's itinerary flow.

8. Verify that the accommodation has the necessary amenities and services required for the traveler's comfort and convenience, such as Wi-Fi if needed for planning or work purposes.

\end{tcolorbox}
\vspace{-3mm}
\caption{An example of the outline}
\label{fig:outline}
\end{wrapfigure}

\clearpage
\newpage

\subsubsection{Strategy Block Content:}
\label{sec:strategy_block}

\begin{wrapfigure}{t}{\textwidth}
\begin{tcolorbox}[halign=left, width=\linewidth, boxrule=3pt, colback=gray!20, colframe=gray!20, left=0pt, right=10pt]
\fontfamily{pcr}\selectfont
\small

Below is a preliminary outline of your trip, which can serve as a reference for collecting information:

The First Day: from Ontario to Honolulu. Exploring Honolulu.

The Second Day: Exploring Honolulu.

The Third Day: from Honolulu to Ontario.

You now need to gather relevant information to specify the travel plan for the First day.

Remember that the total daily expenses of your trip (the sum of expenses for each person) do not exceed 1066.

Hard Constraints. Total Budget: The trip must not exceed a total cost of \$3,200, including all transportation, accommodation, meals, and activities.

\{Detail Omit\}

Thought 1: To begin planning your first day, I will collect information on a flight from Ontario to Honolulu for March 4th, 2022. After that, I will look for dining options, attractions to visit, and accommodation in Honolulu. Let's start with the flight information.

Action 1: FlightSearch[Ontario, Honolulu, 2022-03-04]

Observation 1: Masked due to limited length. Make sure the data has been written in Notebook. Successfully recorded in Notebook: Flight Information for Ontario to Honolulu on March 4, 2022

Thought 2:...

\{Omit Other Process\}

Action 5: DailyPlanner[Please outline the plan for the first day of a 3-day trip from Ontario to Honolulu on March 4th, 2022, including flight, dining, attractions, and accommodation.]
Observation 5: Travel Plan:

Day 1:

Current City: from Ontario to Honolulu

Transportation: Flight Number: F3584294, from Ontario to Honolulu, Departure Time: 10:11, Arrival Time: 14:46

Breakfast: - (Assuming breakfast is eaten before departure or on the flight)

Attraction 1: Aloha Tower, Honolulu (Estimated visit time: 1 hour)

Lunch: Bait El Khetyar, Honolulu (Cuisines: Cafe, Seafood, Average Cost: 28)

Attraction 2: Iolani Palace, Honolulu (Estimated visit time: 1.5 hours)

Attraction 3: Honolulu Museum of Art (HoMA), Honolulu (Estimated visit time: 2 hours)

Dinner: Crystal Restaurant, Honolulu (Cuisines: Tea, Pizza, Bakery, Fast Food, Chinese, Average Cost: 57)

Accommodation: Park, Subway \& All Conveniences; Room type: Entire home/apt; Price: 754; City: Honolulu

Outline:

The First Day: from Ontario to Honolulu. Exploring Honolulu.

The Second Day: Exploring Honolulu.

The Third Day: from Honolulu to Ontario.

Remember that the total daily expenses of your trip (the sum of expenses for each person) do not exceed 1066.

You should gather the necessary information to plan your trip for the Second day.

Each time you use the DailyPlanner tool, the planner can only access information queried during the previous 2 days. However, if the number of queries in the past 2 days is less than 5, then it will return the last 5 queried pieces of information.

Thought 6: ...

\{Omit Future Steps\}

\end{tcolorbox}
\vspace{-3mm}
\caption{An example of the content of strategy block}
\label{fig:strategy_block}
\end{wrapfigure}

\clearpage
\newpage
\subsubsection{Knowledge Block Pop out Content:}
\label{sec:knoledge_block_format}

\begin{wrapfigure}{h}{0.9\textwidth}
\begin{tcolorbox}[halign=left, width=\linewidth, boxrule=3pt, colback=gray!20, colframe=gray!20, left=0pt, right=10pt]
\fontfamily{pcr}\selectfont
1: Flight Number: F3502691; Price: 240; DepTime: 18:48; ArrTime: 20:51; OriginCityName: Buffalo; DestCityName: Atlanta

2: Flight Number: F3514187; Price: 322; DepTime: 06:51; ArrTime: 08:40; OriginCityName: Buffalo; DestCityName: Atlanta

3: Flight Number: F3555201; Price: 265; DepTime: 12:44; ArrTime: 14:33; OriginCityName: Buffalo; DestCityName: Atlanta

...

-------------------------------------

1: Attraction Name: Atlanta Botanical Garden; City: Atlanta

2: Attraction Name: World of Coca-Cola; City: Atlanta

3: Attraction Name: Georgia Aquarium; City: Atlanta

...

--------------------------------------

1: Restaurant: Adda; City: Atlanta; Cuisines: Tea, Mexican, Bakery; Average Cost: 22; Rating: 3.8

2: Restaurant: Baba Au Rhum; City: Atlanta; Cuisines: Desserts, Pizza, Mexican, BBQ, Fast Food; Average Cost: 27; Rating: 4.5

3: Restaurant: Barkat; City: Atlanta; Cuisines: Bakery, Indian, Mediterranean, Desserts; Average Cost: 78; Rating: 3.4

...

--------------------------------------

1: Accommodation: Fantastic Room in Bushwick; Room type: Private room; Price: 1069.0; Minimum number of nights stay: 2.0; review rate number: 3.0; House rules: No children under 10; One room can accommodate how many people: 2; City: Atlanta

2: Accommodation: Sunny, Friendly, Brooklyn Apartment; Room type: Private room; Price: 874.0; Minimum number of nights stay: 1.0; review rate number: 4.0; House rules: No pets; One room can accommodate how many people: 2; City: Atlanta

3: Accommodation: 1bd in a sunny 2 bd Ft. Greene Apt; Room type: Private room; Price: 1056.0; Minimum number of nights stay: 1.0; review rate number: 4.0; House rules: No visitors \& No pets; One room can accommodate how many people: 1; City: Atlanta

...
\end{tcolorbox}

\caption{An example of format information in Knowledge Block.}
\label{sec:knowledge_block}
\end{wrapfigure}
\clearpage

\subsection{Prompt}

\subsubsection{PathFinder Agent Prompt:}
\begin{figure}[htbp]
    \centering
    \includegraphics[page=1, width=\textwidth]{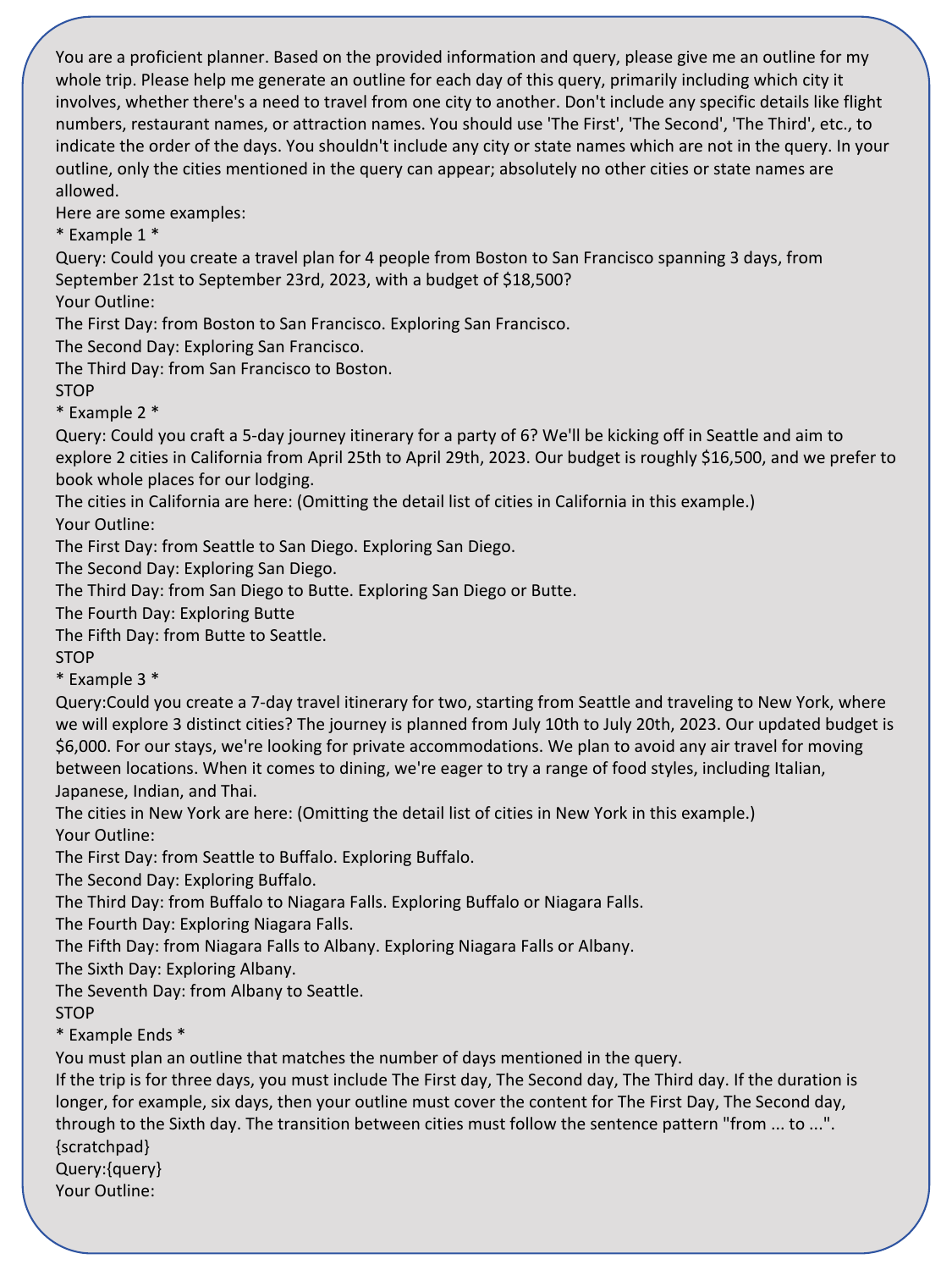}
    \vspace{-4mm}
    \caption{PathFinder Agent Prompt}
    \label{fig:pathfinder_prompt}
    \vspace{-6mm}
\end{figure}

\clearpage

\newpage
\subsubsection{Thought Agent Prompt:}

\begin{figure}[htbp]
    \centering
    \includegraphics[page=2, width=\textwidth]{img/prompt.pdf}
    \vspace{-4mm}
    \caption{Thought Agent Prompt}
    \label{fig:thought_agent_prompt}
    \vspace{-6mm}
\end{figure}
\clearpage
\newpage

\subsubsection{Tool Agent Prompt:}
\begin{figure}[htbp]
    \centering
    \includegraphics[page=3, width=\textwidth]{img/prompt.pdf}
    \vspace{-4mm}
    \caption{Tool Agent Prompt}
    \label{fig:tool_agent_prompt}
    \vspace{-6mm}
\end{figure}
\clearpage
\newpage

\subsubsection{Plan Agent Prompt:}

\begin{figure}[htbp]
    \centering
    \includegraphics[page=4, width=\textwidth]{img/prompt.pdf}
    \vspace{-4mm}
    \caption{Plan Agent Prompt}
    \label{fig:plan_agent_prompt}
    \vspace{-6mm}
\end{figure}
\clearpage
\newpage

\end{document}